\documentclass[sigconf]{acmart}

\AtBeginDocument{%
  \providecommand\BibTeX{{%
    \normalfont B\kern-0.5em{\scshape i\kern-0.25em b}\kern-0.8em\TeX}}}

\copyrightyear{2025}
\acmYear{2025}
\setcopyright{rightsretained}
\acmConference[RecSys '25]{Nineteenth ACM Conference on Recommender Systems}{September 22--26, 2025}{Prague, Czech Republic}
\acmBooktitle{Nineteenth ACM Conference on Recommender Systems (RecSys '25), September 22--26, 2025, Prague, Czech Republic}

\newif\ifproofread

\usepackage{color,soul}
\usepackage{pgfplots}
\usepackage{hyperref}
\usepackage{framed}
\usepackage{fontawesome}

\usepackage{bm}
\usepackage{multirow}
\usepackage{makecell}
\usepackage{booktabs} 
\usepackage{multirow,bigdelim}
\usepackage{easybmat}
\usepackage{threeparttable}

\usepackage{pgfplotstable}
\pgfplotsset{compat=newest}
\usepackage{caption}
\usepackage{subcaption}
\usepackage{tikz}

\usepackage{xspace}
\usepackage{xcolor}
\usepackage{xargs} 
\usepackage[colorinlistoftodos,prependcaption,textsize=tiny]{todonotes}

\newcommandx{\add}[2][1=]{\todo[linecolor=red,backgroundcolor=red!25,bordercolor=red,#1]{#2}}
\newcommandx{\change}[2][1=]{\todo[linecolor=blue,backgroundcolor=blue!25,bordercolor=blue,#1]{#2}}
\newcommandx{\info}[2][1=]{\todo[linecolor=OliveGreen,backgroundcolor=OliveGreen!25,bordercolor=OliveGreen,#1]{#2}}
\newcommandx{\improvement}[2][1=]{\todo[linecolor=orange,backgroundcolor=orange!25,bordercolor=orange,#1]{#2}}

\usepackage{listings}
\usepackage{xcolor}

\definecolor{codegreen}{rgb}{0,0.6,0}
\definecolor{codegray}{rgb}{0.5,0.5,0.5}
\definecolor{codepurple}{rgb}{0.58,0,0.82}
\definecolor{backcolour}{rgb}{0.95,0.95,0.92}

\lstdefinestyle{mystyle}{
    backgroundcolor=\color{backcolour},   
    commentstyle=\color{codegreen},
    keywordstyle=\color{magenta},
    numberstyle=\tiny\color{codegray},
    stringstyle=\color{codepurple},
    basicstyle=\ttfamily\footnotesize,
    breakatwhitespace=false,         
    breaklines=false,                 
    captionpos=b,                    
    keepspaces=false,                 
    numbers=left,                    
    numbersep=4pt,                  
    showspaces=false,                
    showstringspaces=false,
    showtabs=false,                  
    tabsize=1
}

\lstdefinestyle{pythonstyle}{
    language=Python,
    backgroundcolor=\color{gray!5},
    commentstyle=\color{green!50!black},
    keywordstyle=\color{blue},
    numberstyle=\tiny\color{gray},
    stringstyle=\color{orange},
    basicstyle=\ttfamily\footnotesize,
    breaklines=true,
    numbers=left,
    numbersep=5pt,
    frame=single,
    rulecolor=\color{gray!40}
}

\usepackage{float}

\usepackage{amsmath,amsfonts,bm}

\def\eqref#1{equation~\ref{#1}}

\def\1{\bm{1}}

\DeclareMathAlphabet{\mathsfit}{\encodingdefault}{\sfdefault}{m}{sl}
\SetMathAlphabet{\mathsfit}{bold}{\encodingdefault}{\sfdefault}{bx}{n}

\usepackage{soul}
\begin{document}

\title{Kamae: Bridging Spark and Keras for Seamless ML Preprocessing}


\author{George Barrowclough}
\affiliation{
    \institution{Expedia Group}
    \city{London, UK}
}
\email{gbarrowclough@expediagroup.com}
\author{Marian Andrecki}
\affiliation{
    \institution{Expedia Group}
    \city{London, UK}
}
\email{mandreki@expediagroup.com}
\author{James Shinner}
\affiliation{
    \institution{Expedia Group}
    \city{London, UK}
}
\email{jshinner@expediagroup.com}
\author{Daniele Donghi}
\affiliation{
    \institution{Expedia Group}
    \city{Geneva, Switzerland}
}
\email{ddonghi@expediagroup.com}

\begin{abstract}
\textbf{Disclaimer:} This is the authors' preprint version of a paper accepted at the 19th ACM Conference on Recommender Systems (RecSys 2025). The final version will appear in the ACM Digital Library.

\vspace{0.5em}
In production recommender systems, feature preprocessing must be faithfully replicated across training and inference environments. This often requires duplicating logic between offline and online environments, increasing engineering effort and introducing risks of dataset shift. We present \textit{Kamae}, an open-source Python library that bridges this gap by translating \textit{Spark} preprocessing pipelines into equivalent \textit{Keras} models. \textit{Kamae} provides a suite of configurable \textit{Spark} transformers and estimators, each mapped to a corresponding \textit{Keras} layer, enabling consistent end-to-end preprocessing across the ML lifecycle. The utility of the framework is illustrated via real-world use-cases, including the MovieLens dataset and Expedia’s Learning-to-Rank pipelines. The code is available at \url{https://github.com/ExpediaGroup/kamae}.

\end{abstract}


\maketitle

\section{Problem Statement}\label{sec:introduction}

Many e-commerce companies, including Expedia (an online travel agent), aim to provide their users with an adaptive and personalised shopping experience. This can be achieved by tracking user-platform interactions and training machine learning models with particular optimisation objectives, such as retention or booking rate. For large-scale vendors, vast amounts of standardised data are stored in a centralised data lake, where it is available to many different teams.

The development of a model starts with the extraction of a segment of standardised data. Generally, the data needs to be transformed before it can be consumed by the model. Useful transformations vary greatly with the data involved, model type, and use-case specifics. For example, this may involve: bringing numerical features to a similar scale, indexing string categorical features to integers, one-hot encoding, creating new features from the existing ones - by regular expression extraction, applying mathematical operators to single or multiple numericals, and many more. The result is a model-specific dataset. We refer to this step as \textit{preprocessing}.

After the model is optimised and evaluated offline, it needs to be deployed in the online inference environment. It is typically available via an API where the client supplies model inputs as part of the request. The client can readily provide features in the standardised data-lake format, but the preprocessed features are generally not available. This implies that the inference needs to start from raw feature values, and the features that result from preprocessing need to be generated within the inference environment. 

There are two main solutions to this problem. (1) The processed features can be calculated by reimplementing the transformations from the preprocessing step. This is limited mostly to features that result from column-wise transformations, as only a single row of data is assumed to be available at inference. (2) Processed features can be retrieved from dedicated tables known as feature stores, which are updated at the desired frequency. This involves building more infrastructure and incurs a retrieval latency penalty.

These approaches are complementary rather than exclusive. In this work, we focus on the first one.

Conceptually, reimplementing preprocessing operations appears trivial. However, in practice, there are a number of engineering challenges, as the model development and inference environments differ significantly. 

The model development environment is relatively unconstrained provided that the final result (the model) can be moved to the inference environment. However, in the context of big data, it is crucial that throughput is high - usually distributed computing is a must.

In the inference environment, the focus is on a low error rate and response time with efficient use of resources. Additionally, companies have preferred ways of serving models, which limits choice of programming languages and libraries.

The code preprocessing transformations may need to be rewritten in another language, often by another specialist with only abstract understanding of the model. This leads to code duplication, increased development and maintenance time, and increased chance of mismatch between the offline and the online models.

\section{Kamae}

\subsection*{Overview}

This work introduces \textit{Kamae},\footnote{\url{https://github.com/ExpediaGroup/kamae}} a novel open-source Python package that bridges the gap between model development and inference environments for data preprocessing purposes. \textit{Kamae} defines a number of configurable \textit{Apache Spark} \cite{zaharia2016apache} transformers and estimators, each mapped to an equivalent \textit{Keras} \cite{chollet2015keras} layer. We extend \textit{Spark}'s pipeline API to enable one-to-one conversion into a \textit{Keras} model. At that point, optionally, the resulting bundle can be fused with the trained model. This allows users to preprocess datasets using \textit{Spark} and seamlessly reflect this computation in deployment by serving a \textit{Keras} model - with a variety of solutions available, e.g. \cite{olston2017tensorflowserve, kserve2021}. At present, only \textit{TensorFlow} \cite{abadi2016tensorflow} backend is supported. Practical usage details such as the full list of available transformations, pipeline examples, and installation guide can be found in the recently open-sourced \href{https://github.com/ExpediaGroup/kamae}{repository} (Apache License 2.0).

\subsection*{Basic Functionalities}

\textit{Kamae} offers a wide range of transformers (mathematical, string, date, geographical, logical, array, list, and conditional operations) and a number of estimators (string-, hash-, bloom-, shared- indexing, standard scaling, and imputation). They are implemented in accordance with the \textit{Spark} API and built using native transformations (rather than user-defined functions) to guarantee high performance. Each transformation contains a method for the construction of the corresponding \textit{Keras} layer. Extensive unit tests ensure parity between \textit{Spark} and \textit{Keras} implementations. We hope open-sourcing the library will extend the list; see section \ref{sec:future}.

Transformations can be chained together, forming a \textit{ pipeline graph}, which is applied (or fitted) to the data in a distributed manner. The graph can then be exported and served as a generic \textit{Keras} model (without \textit{Kamae}'s package dependencies). See the examples in Section \ref{sec:use_cases}.

\subsection*{Advanced Functionalities}

\textbf{Indexing}. Beyond usual string indexing that maps strings to integers via vocabulary lookup, there are a few more configurable options. Shared string indexing allows for vocabulary to be built and applied across multiple features. Hash indexing enables mapping to an integer when the cardinality of categories is overwhelming. Bloom encoding \cite{serra2017getting} is a technique that applies hash indexing multiple times allowing for memory-efficient encoding of high-cardinality categoricals.

\textbf{Nested-sequence-native}. \textit{Kamae} has been developed by a team that focuses on Learning-to-Rank problems. In this setting, sequence features are frequently encountered. For example, in Expedia's recommendation flows, users are often presented with ranked lists of options (e.g.  hotels or rooms), and each item may carry structured features such as a list of amenities. \textit{Kamae} offers transformations that can be applied element-wise (preserving the dimensionality) or applied at the sequence level (aggregating or transforming the list as a whole). For instance, string indexing can be applied independently to each element of a list-type feature (e.g., amenities), yielding a padded integer sequence.

\textbf{Keras Tuner support}. Optionally, an exported preprocessing model can be fused with a neural model to be trained. At that point, \textit{Keras Tuner} \cite{omalley2019kerastuner} can be configured to search for the best hyperparameter settings of the preprocessing layers. This is particularly useful for tuning parameters such as the number of hash bins, embedding dimensions, or thresholds in feature engineering steps. By integrating preprocessing into the hyperparameter optimization process, practitioners can systematically explore and identify configurations that yield optimal model performance.

\subsection*{Similarity to Existing Frameworks}

The closest existing solution is \textit{MLeap} \cite{mleap}, built in Scala. It offers a wide range of \textit{Spark} transformations implemented as user-defined functions, which prevents the execution engine from optimising the operations with \textit{Catalyst}\cite{armbrust2015spark}. The resulting pipeline can be serialised to a custom \textit{Bundle.ML} format. At inference time, \textit{MLeap Runtime} along with \textit{Java Virtual Machine} (JVM) are required to execute the model.

The key distinct features of \textit{Kamae} include:
(1) Built with Python and does not require \textit{JVM} at inference.
(2) The serialised model is a \textit{Keras} bundle - a popular standard that enables performant serving.
(3) Native support for (nested) array-type features.

Scikit-learn pipelines are conceptually similar to our approach. These allow for a number of high-level ML-centric operations (scaling, encoding, dimensionality reduction, and more) to be composed into a single transformer, which can then be conveniently deployed in a Python environment. However, this solution does not allow for the use of distributed compute, so it cannot be applied to large datasets.

\section{Use-case examples}\label{sec:use_cases}

\subsection*{MovieLens Pipeline}

For illustrative purposes, in Listing \ref{list:kamae_pipeline} we provide code for a minimal pipeline to transform the MovieLens dataset\cite{harper2015movielens}. As the pipeline contains estimators, it has to be fit to the data. Then it can be exported to a \textit{Keras} model, which would provide equivalent transformations at inference.

\begin{figure}[ht]
  \centering
  \captionsetup{type=figure}
  \begin{minipage}{\linewidth}
    \begin{lstlisting}[style=pythonstyle, caption={Kamae preprocessing pipeline for the MovieLens dataset}, label={list:kamae_pipeline}]
user_hash_indexer = HashIndexTransformer(
    inputCol="UserID",
    outputCol="UserID_indexed",
    # Set the inputDtype to force the id to be a string
    inputDtype="string",
    # Set 10k bins to reduce collisions
    numBins=10000,
    layerName="user_hash_indexer",
)
movie_id_string_indexer = StringIndexEstimator(
    inputCol="MovieID",
    outputCol="MovieID_indexed",
    inputDtype="string",
    # Order the collected labels by descending frequency
    stringOrderType="frequencyDesc",
    numOOVIndices=1,
    layerName="movie_id_string_indexer",
)
occupation_one_hot_encoder = OneHotEncodeEstimator(
    inputCol="Occupation",
    outputCol="Occupation_indexed",
    stringOrderType="frequencyDesc",
    inputDtype="string",
    numOOVIndices=1,
    # Whether the one hot encoder should drop the index for unseen. 
    dropUnseen=True,
    layerName="occupation_one_hot_encoder",
)
genres_split_to_array_transform = StringToStringListTransformer(
    inputCol="Genres",
    outputCol="Genres_split",
    separator="|",
    # Max number of genres for a movie is 6
    listLength=6,
    # If a list does not have 6 it will be padded
    defaultValue="PADDED",
    layerName="genres_split_to_array_transform",
)
genres_string_indexer = StringIndexEstimator(
    # Input is the output of the splitting
    inputCol="Genres_split",
    outputCol="Genres_indexed",
    stringOrderType="frequencyDesc",
    numOOVIndices=1,
    # Mask the PADDED token to send this to the 0 index
    maskToken="PADDED",
    layerName="genres_string_indexer",
)
pipeline = KamaeSparkPipeline(
    stages=[
        user_hash_indexer,
        movie_id_string_indexer,
        occupation_one_hot_encoder,
        genres_split_to_array_transform,
        genres_string_indexer,
    ]
)
fit_pipeline = pipeline.fit(train_ml)
input_schema = [
    dict(name="UserID", dtype="int32", shape=(1,)),
    dict(name="MovieID", dtype="int32", shape=(1,)),
    dict(name="Occupation", dtype="int32", shape=(1,)),
    dict(name="Genres", dtype="string", shape=(1,)),
]
keras_model = fit_pipeline.build_keras_model(tf_input_schema=input_schema)
    \end{lstlisting}
  \end{minipage}
\end{figure}

\subsection*{Learning-to-Rank Search Filters}

Expedia employs Learning-to-Rank models to personalise hotel search experiences, tailoring results based on user preferences and destination-specific popularities, and more. A notable application is the ranking of search filters (e.g., amenities such as pools or spas) to enhance user engagement and booking rates.

The simplified model development flow with particular focus on \textit{Kamae} pipelining begins with preprocessing of the dataset. Initially, invalid rows/queries are removed from the dataset. Subsequently, \textit{Kamae} transformations are configured to instantiate a \textit{Spark} pipeline.

In this practical case, the list of transformations is extensive and strongly tied to the number of features used. We summarise it briefly.

\begin{itemize}
    \item date features are disassembled into parts, e.g. month, weekday, so that the model can accommodate for seasonality
    \item particular dates are subtracted to generate durations
    \item numerical values spanning many orders of magnitude are log-transformed
    \item particular string-valued features are split into multiple new features based on delimiters
    \item selected numerical features are assembled into a single array which is subsequently standard scaled and disassembled into original features
    \item categorical features are indexed
    \item overall, this involves around 60 transforms, often chained together
\end{itemize}

The pipeline is then fitted and applied to the data, resulting in a transformed dataset suitable for training and evaluation.

Once the model is trained, evaluated, and selected for deployment, it can be fused with the \textit{Kamae} pipeline. In the production environment, this fused model bundle is loaded within a Java chassis. The model is executed on client request with average rate of 200 requests per second.

Previously, this workflow relied on \textit{MLeap}. The transition to a \textit{Keras} model and the use of \textit{TensorFlow Java} led to a 61\% decrease in service latency and a 58\% reduction in service costs. Further improvements are expected after migration to \textit{KServe} \cite{kserve2021}.

Beyond quantitative enhancements, the project codebase has been simplified by replacing the Scala preprocessing stage with Python code, effectively removing the Scala dependency.

\section{Future Work}\label{sec:future}

\textit{Kamae} is released as an open source library to encourage adoption, feedback, and contributions from the broader machine learning and recommender systems communities. We believe that interoperability between training and inference environments is a shared challenge and invite collaboration to evolve the library beyond its current scope.

In future work, we plan to extend \textit{Kamae} along three primary axes:

\textbf{Backend support beyond TensorFlow:} While the current implementation targets \textit{Keras} models for \textit{TensorFlow} inference, we aim to support additional backends such as \textit{JAX} and \textit{PyTorch}.

\textbf{Expanded transformation coverage:} We intend to grow the library of available transformations, focussing on both commonly used preprocessing steps (e.g. tokenization, quantile binning) and more complex feature engineering primitives that capture domain-specific signals.

\textbf{Community-driven enhancements:} As an open-source project, \textit{Kamae} is structured to enable external contributions. We plan to incorporate community feature requests, production use case extensions, and performance optimisations based on real-world feedback.

\section{Author Biographies}\label{sec:bios}

\textbf{George Barrowclough} is a scientist in the Relevance Team at Expedia Group. He holds a Masters in Mathematics from the University of Oxford.
\\
\textbf{Marian Andrecki} is a scientist in the Relevance team at Expedia Group. He obtained his PhD in Machine Learning from Heriot-Watt and Edinburgh Universities.
\\
\textbf{James Shinner} is a scientist in the Relevance team at Expedia Group. He obtained his PhD in Elementary Particle Physics from Royal Holloway, University of London and CERN.
\\
\textbf{Daniele Donghi} is a senior scientist in the Ranking team at Expedia Group. He holds a Master in Compute Science from the University of Illinois Chicago (UIC), and a Master in Compute Engineering from Polytechnic University of Milan.

\section*{Acknowledgments}

The authors thank the contributors at Expedia Group; their help has been invaluable in developing and refining Kamae. We appreciate the valuable feedback on this paper from Andrea Marchini, Victoria Lopez and Giorgio Mondauto.

\bibliographystyle{ACM-Reference-Format}
\bibliography{biblio}

\end{document}